\documentclass[sigconf]{acmart}

\AtBeginDocument{%
  \providecommand\BibTeX{{%
    \normalfont B\kern-0.5em{\scshape i\kern-0.25em b}\kern-0.8em\TeX}}}

\setcopyright{rightsretained}
\copyrightyear{2019}
\acmYear{2019}
\acmDOI{}

\acmConference[KDD Workshop on AI for fashion]{KDD}{August, 2019}{Anchorage, AK}
\acmPrice{}
\acmISBN{}

\begin{document}

\title{Pose Guided Fashion Image Synthesis Using Deep Generative Model}

\author{Wei Sun}
\authornote{The work was done when Wei Sun and Yi Xu were with JD.com American Technologies.}
\affiliation{%
  \institution{North Carolina State University}
  \city{Raleigh}
  \state{NC}
  \country{USA}
}
\email{wsun12@ncsu.edu}

\author{Jawadul H. Bappy}
\affiliation{%
 \institution{JD.com American Technologies}
   \city{Mountain View}
   \state{CA}
  \country{USA}
}
\email{jawadul.bappy@jd.com }

\author{Shanglin Yang}
\affiliation{%
  \institution{JD.com American Technologies}
     \city{Mountain View}
   \state{CA}
  \country{USA}
}
\email{shanglin.yang@jd.com}

\author{Yi Xu}
\authornotemark[1]
\affiliation{%
  \institution{OPPO US Research Center}
     \city{Palo Alto}
   \state{CA}
  \country{USA}
}
\email{yi.xu@oppo.com}

\author{Tianfu Wu}
\affiliation{
\institution{North Carolina State University}
   \city{Raleigh}
   \state{NC}
  \country{USA}}
  \email{tianfu_wu@ncsu.edu}

\author{Hui Zhou}
\affiliation{
\institution{JD.com American Technologies}
   \city{Mountain View}
   \state{CA}
  \country{USA}}
\email{hui.zhou@jd.com}


\begin{abstract}
Generating a photorealistic image with intended human pose is a promising yet challenging research topic for many applications such as smart photo editing, movie making, virtual try-on, and fashion display.
In this paper, we present a novel deep generative model to transfer an image of a person from a given pose to a new pose while keeping fashion item consistent. In order to formulate the framework, we employ one generator and two discriminators for image synthesis. The generator includes an image encoder, a pose encoder and a decoder. The two encoders provide good representation of visual and geometrical context which will be utilized by the decoder in order to generate a photorealistic image. Unlike existing pose-guided image generation models, we exploit two discriminators to guide the synthesis process where one discriminator differentiates between generated image and real images (training samples), and another discriminator verifies the consistency of appearance between a target pose and a generated image. We perform end-to-end training of the network to learn the parameters through back-propagation given ground-truth images. The proposed generative model is capable of synthesizing a photorealistic image of a person given a target pose. We have demonstrated our results by conducting rigorous experiments on two data sets, both  quantitatively and qualitatively.
\vspace{1mm}
\end{abstract}





\maketitle

\section{Introduction}

\begin{figure}[t]
\begin{center}
\begin{tabular}{c}
  \includegraphics[width=.8\linewidth]{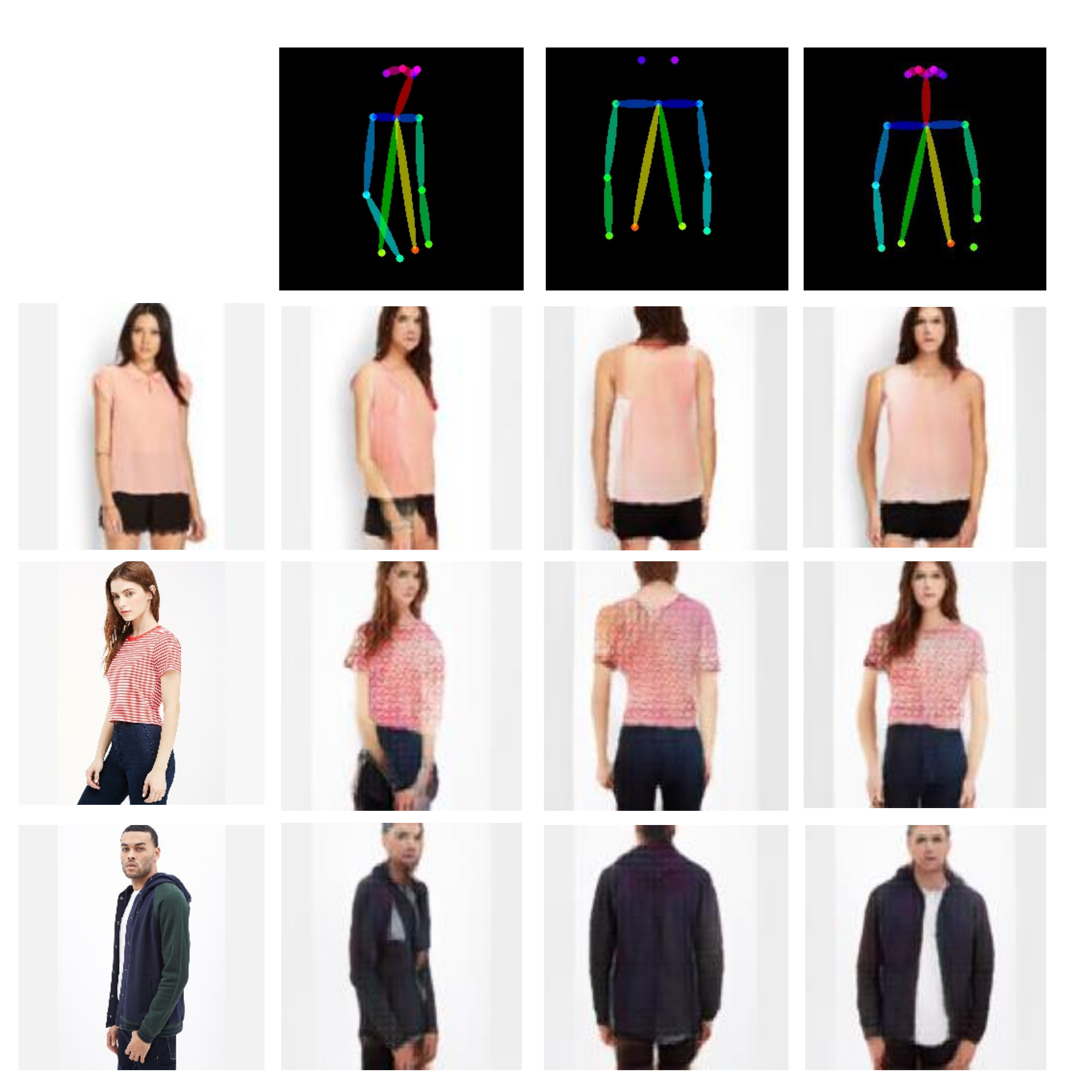} 
 \end{tabular}
 
\end{center}
  \caption{\small Some examples of generated images produced by our proposed model. First column represents the input images, and top row shows the target poses. Given an input image of a person and a target pose, the proposed model is able to replace a person's current pose with the target pose. Best viewable in color.
  }
\label{prob_def} 
\end{figure}
Over the past few years, online fashion industry has been shaped by recent
technological innovations such as augmented reality, virtual reality, wearable tech, and connected fitting rooms. 
In order to attract online shoppers and to deliver rich and intuitive online experience, retailers strive to provide high-quality and informative pictures of the products.   
Online shoppers usually expect to see multiple photos of a garment item from different viewpoints, or multiple photos of a fashion model wearing the same garment from different angles or under different poses. In such scenarios, image synthesis techniques can be exploited to enhance shopping experience for shoppers and to reduce cost for retailers. 
In computer vision, image generative models \cite{goodfellow2014generative,yoo2016pixel,van2016conditional,van2016wavenet}, which are capable of generating high quality photorealistic images, have been successfully applied in numerous applications. 
In this paper, our main objective is to develop an image generative model in order to transfer a person from its current pose to an intended target pose.

Generative Adversarial Network (GAN) \cite{goodfellow2014generative} is one of the prominent approaches for image synthesis that has been widely used. For fashion applications, there have been some prior works 
that utilize generative models at conditional settings.  In \cite{ma2017pose}, a reference image has been utilized to transfer a person with given pose to intended pose. Shape information has been incorporated in \cite{esser2018variational} to aid image generation process. Unlike these two methods, which use one discriminator for pose guided image generation task, we utilize \textit{two} specific discriminators - one discriminator differentiates between real image and generated image, and another one enhances the consistency between the generated image and the target pose. For virtual try-on, Han et al. propose a VITON network \cite{han2018viton}  that virtually dresses a person with an different fashion item. The objective is different between VITON and our work-VITON allows for a user to virtually try on different garments, while our work allows for a online retailer to easily generate various display photos. Moreover, online retailers usually provide multiple photos. In such scenario, it will be advantageous to utilize multiple photos as input in order to extract visual-semantic features both for training and image generation. Unlike most of the image generation approaches \cite{han2018viton,ma2017pose}, we exploit a set of images of the same fashion item, either garment itself or a fashion model wearing the garment, from which a meaningful representation is learned. 

In this paper, we aim to develop a novel generative model to produce  photorealistic images of a person with new pose different from its current pose. 
The proposed framework exploits a bi-directional convolutional LSTM \cite{xingjian2015convolutional,donahue2015long} network and U-Net architecture \cite{ronneberger2015u} for image generation process. 
The LSTM network is utilized to discover the common attributes from multiple images
by observing the change in the various semantic image features, such as colors, textures, and shapes. The network is also capable of distinguishing background or noise from the variation in semantic features.
A U-Net encoder is used to learn a compact representation of appearance. The representations learned from convolutional LSTM and U-Net encoder are then exploited to synthesize a new image. Two discriminators are designed and deployed in order to guide the image generation process. 
We perform end-to-end training of generator and discriminator networks. We show both quantitative and qualitative analysis to evaluate the performance of our image generative model on two datasets.

{\bf Main Contributions.} Our major contributions are as follows.

\begin{itemize}
    \item In this paper, we present a novel generative model which employs two encoders (${\it E}_{\mathcal{I}}$ and ${\it E}_{\mathcal{P}}$) and one decoder to generate a new image. The representations learned by the two encoders from multiple images of the same fashion item is compact and meaningful, which can be applied in other tasks such as image search and garment parsing. 

    \item  The proposed framework exploits two discriminators where one discriminator enforces photorealism of the generated images, and the other discriminator enhances the consistency between generated image and target pose.

    \item Using multiple images (e.g., images of a person wearing same garment item with different poses) allows the convolutional LSTM network to learn more visual-semantic context that helps guide the image generation process.
\end{itemize}



\section{Related Works}

Recently, image generative modeling has gained a lot of attention from both scientific communities and fashion industry. 
Generative Adversarial Networks (GANs) \cite{goodfellow2014generative} are the most popular generative models for the tasks of image synthesis and image modification. There have been some works \cite{van2016conditional,isola2017image} that exploit GAN in conditional setting. In \cite{denton2015deep,van2016conditional}, generative models are developed conditioning upon class labels. Text\cite{reed2016generative,zhu2017your} and images \cite{ma2017pose,isola2017image,yoo2016pixel,wang2018high,lassner2017generative} have also been used as conditions to build image generative models. 


{\bf Computer Vision in Fashion.} 
Recent state-of-the-art approaches demonstrate promising performance in a few computer vision tasks such as object detection \cite{ren2015faster,redmon2016you}, semantic segmentation \cite{long2015fully}, pose estimation \cite{papandreou2017towards}, and image synthesis \cite{goodfellow2014generative,esser2018variational}. Recent state-of-the-art approaches has applied in fashion related tasks, for instance, visual search \cite{yang2017visual}, cloth parsing or segmentation \cite{yamaguchi2015retrieving,tangseng2017looking}, et al. In \cite{hadi2015buy}, the authors present a deep learning based matching algorithm to solve street-to-shop problem.  In \cite{liu2015matching}, parametric matching convolutional neural network (M-CNN) and non-parametric KNN approaches have been proposed for human parsing. In \cite{liu2015fashion}, pose estimation and fashion parsing are performed where SIFT flow and super-pixel matching are used to find correspondences across frames. In garment retrieval task,  fine-grained attribute prediction\cite{wei2013style}, parsing \cite{yamaguchi2012parsing}, and cross-scenario retrieval \cite{fu2012efficient} have been utilized in order to improve performance. Most of the approaches \cite{yoo2016pixel, ma2017pose, lassner2017generative, zhu2017your} exploit deep learning based encoder to generate new images due to its superior performance. In \cite{yoo2016pixel}, an image-conditional image generation model has been proposed to transfer an input domain to a target domain in semantic level.  To generate images of the fashion model with new poses, there have been a few previous efforts. An image synthesis technique conditioned upon text has been portrayed in \cite{yoo2016pixel}. 

{\bf Image Synthesis in Fashion.}
These techniques have also been applied \cite{han2018viton,zhu2017your,ma2017pose} to exploit image generative models in fashion technology. An image based virtual try-on network has been proposed in \cite{han2018viton} where the generative model transfers a desired clothing item onto the corresponding region of a person using a coarse-to-fine strategy. A novel approach is presented in \cite{zhu2017your} for generating new clothing on a wearer through generative adversarial learning by utilizing textual information. In \cite{esser2018variational}, a conditional U-Net has been used to generate image guided by shape information, and conditioned on the output of a variational autoencoder  for appearance. In \cite{ma2017pose}, the authors present a generative model conditioned upon pose to manipulate a person in an image to an arbitrary pose. \cite{chan2018everybody} study similar task with us, instead they transfer poses to target person from video in a frame-by-frame manner.

Even though we aim at solving similar problem as \cite{ma2017pose}, our work differs from \cite{ma2017pose} in terms of architectural choices both in generator and discriminator. Unlike most of the image generative approaches, we exploit  multiple images of a fashion item as input which are usually available on e-commerce shopping platforms.

\section{Proposed Model}
\label{DeepModel}

\begin{figure}[h]
\begin{center}
  \includegraphics[width=1.0\linewidth]{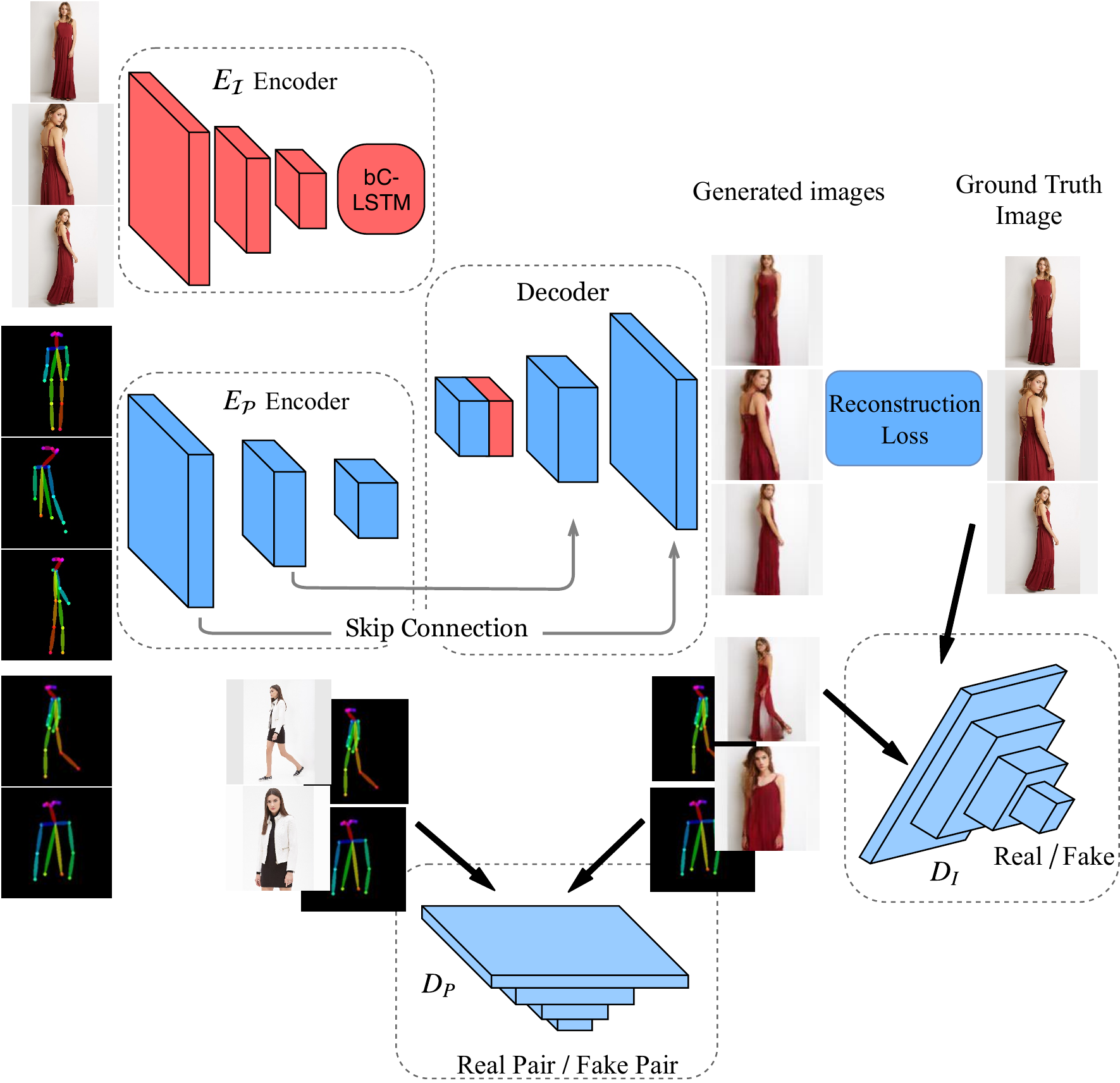}
  \end{center}
  \caption{\small The figure presents the overview of our image generative framework. In this figure, ${\it E}_{\mathcal{I}}$ and ${\it E}_{\mathcal{P}}$ encoders exploit bi-directional convolutional LSTM network and U-Net encoder respectively. The fusion of encoded features from both encoders are utilized by U-Net decoder to synthesize an image. During training phase, two discriminators $D_{I}$ and $D_{P}$ have been utilized in order to generate the photo-realistic image of a person while maintaining shape consistency with an intended pose. Best viewable in color.}
  \label{proposed_model}
\end{figure}

Given a set of images and pose maps as input, our objective is to generate photorealistic image of a person with a new pose different from current one.
The proposed framework has two basic components - (a) generator, and (b) discriminator. 
Fig.~\ref{proposed_model} demonstrates the overall architecture.

\subsection{Generator}
In this paper, we develop a generator $\mathcal{G}$ to produce a photorealistic image of a person with a target pose. Our 
generator has  three parts: (a) ${\it E}_\mathcal{I}$ encoder, (b) ${\it E}_\mathcal{P}$ encoder, and (c) a decoder. 
Fig.~\ref{proposed_model} illustrates how different components are utilized to form an image generator. The generator exploits visual-semantic context and pose information obtained from ${\it E}_\mathcal{I}$ and ${\it E}_\mathcal{P}$ encoder respectively, which will then be fed into a decoder in order to generate a new image.

\subsubsection{Image Encoder ${\it E}_\mathcal{I}$} 
The objective of  ${\it E}_\mathcal{I}$ is to learn a semantic representation  from a set of images or from a single image. 
To extract visual features from images, we use ResNet \cite{he2016deep} architecture that includes several residual blocks. At each layer, the network learn different features, e.g., texture, color, edges, contours. Next, these features are fed to a bidirectional convolutional LSTM (bC-LSTM). 
While LSTM has been used in several recognition tasks to extract  sequential information. The main motivation of using the bC-LSTM network in our work is to connect the  common features from the same person wearing same fashion item at different viewpoints.
The bC-LSTM network 
observes the transition of various semantic image features, such as colors, textures, and shapes from one image to other. As a result, the network can also distinguish background and noise from the variation in features.

After training, ${\it E}_\mathcal{I}$ is able to learn
useful visual-semantic  representation of the images.    The learned representation or `codes' match the concepts of different aspects of fashion items, such as semantic components of the garment (e.g., sleeves, neck, etc.) and certain textural information of the fashion item. 
We denote the representation as $C_I$.
The representative code ${C}_I$ learned from ${\it E}_\mathcal{I}$ will be utilized by a decoder to generate new image.

 
\subsubsection{Pose Encoder ${\it E}_{\mathcal{P}}$}
Fig.~\ref{proposed_model} also shows the ${\it E}_{\mathcal{P}}$ encoder used in our framework. 
We use
a U-Net architecture \cite{ronneberger2015u} to encode the pose information. 
We provide pose feature maps with $3$ channels (R, G, B) as input to the network. Human pose estimation method \cite{cao2017realtime} is used to generate locations of $18$ keypoints. We create a pose map  by joining the keypoints with a straight line using different colors as shown in Fig.~\ref{proposed_model}.

The map will then be used by U-Net encoder to aggregate geometrical features. The U-Net  encoder includes two $3\times 3$ convolutions. Each convolution is followed by a rectified linear unit (ReLU) and a $2\times 2$ max pooling operations. We also increase the number of feature channels by $2$ as in \cite{ronneberger2015u}. Each layer of U-Net  encoder is connected to later layers of U-Net decoder by skip connections in order to produce high level features. Finally, we obtain a representation $C_P$. In the following section, we will discuss how the outputs of  ${\it E}_{\mathcal{I}}$ encoder and  ${\it E}_{\mathcal{P}}$ encoder have been further utilized in the decoder network.

\subsubsection{Decoder}
The primary focus of the decoder is to generate a new image by decoding the representative codes ${C}_{{I}}$ and $
{C}_{{P}}$ obtained from $E_\mathcal{I}$ and $E_\mathcal{P}$ encoders respectively.  
The encoded features $C_I$ and $
C_P$ are concatenated in the intermediate stage which will be taken as input to Decoder. Fig.~\ref{proposed_model} shows the steps of image synthesis process. 
For decoder, we use convolutional decoder from U-Net architecture with skip connections. 
The advantage of using the skip connections with ${\it E}_{\mathcal{P}}$ encoder is that it allows the network to align the visual-semantic features with appearance context learned in U-Net encoder.

We fuse the visual ${C}_{{I}}$ and pose ${C}_{P}$ encoded features computed from ${E}_\mathcal{I}$ and ${E}_\mathcal{P}$ respectively. The fused feature maps are fed to the U-Net decoder. At each layer of decoder, we first aggregate the feature maps obtained from previous layer and the precomputed feature maps at early stage chained by skip connection. Next, we upsample the feature map which is followed by $2\times 2$ up-convolution. This operation also decreases the number of channels by half. Up-convolution is followed by $3\times 3$ convolution and ReLU operation. Finally, we obtain a synthesized image $\hat{{y}}$ as output of U-Net decoder. 

\subsection{Discriminator}
The main objective of discriminator is to guide the image generation process to  be photorealistic by comparing synthesized images against genuine ones. 
During the training process of the network, we apply two discriminators: discriminator $D_{I}$ classifying whether an image is real or fake (generated); and discriminator $D_{P}$
aiming to estimate whether a pair,  e.g., an image of a person and a pose, is consistent.
The architectures for $D_{I}$ and $D_{P}$ are shown in bottom right of Fig.~\ref{proposed_model}.

Similar to other traditional GAN models, we use a discriminator network $D_{I}$ to guide the generation of an image. The $D_{I}$ discriminator  distinguishes between an real image and fake (generated) image. 
Sometimes, the generated images looks `real', but not consistent with pose provided. In this paper, we propose another discriminator, denoted as $D_P$, which aims to distinguish between a generated image-pose pair $(\hat{{y}}, p)$ and a real image-pose pair $({{y}}, p)$ by checking the consistency between them. Here, $\hat{{y}}, {y}$ and $p$ represent the real, generated (fake) and pose map.
This discriminator plays a vital role to align a person with a target pose.  Thus, our model can also generate images with complicated pose by enforcing consistency. Exploitation of two discriminators makes our image generation process more robust, consistent and photorealistic.

\subsection{Training}
 
During the training of generator, we define the loss function in a way so that the generated image is judged as `real' and `consistent' with corresponding pose provided by the discriminators. 
In contrast, the loss functions for discriminators are chosen to predict the newly generated image as fake or inconsistent with high confidence. 
We take advantage of the adversarial examples to train the whole network simultaneously.
After optimization of the parameters, the proposed generator is able to generate photorealistic images similar to the training images which cannot be distinguished from real images by the two discriminators. 

Let us denote a set of images that belong to a same person wearing same fashion garment with different pose as $\{x_{i}\}_{i=1}^N$, and $\{p_{i}\}_{i=1}^N$ represent the corresponding pose maps \footnote{For simplicity, we often omit the subscript.}, where $N$ is the number of images. 
The generator $G$ generates a set of images $\{\hat{y}_{i}\}_{i=1}^N$ given $\{x_{i}\}_{i=1}^N$, and $\{p_{i}\}_{i=1}^N$. Here, $G$ indicates a combination of Image Encoder, Pose Encoder, and Decoder. The
generator model $G$ learns a mapping function $ G(x_i, p_i) = \hat{y}_i$. 
Using the ground-truth images, we can write the loss function for the generator as
\begin{equation}
 L_{{G}}(G) =||y - G(x, p) ||_1  + \sum\limits_{k} \lambda_k || \Phi_k(y) - \Phi_k(G({x , p})) ||_1
 \label{loss_gen}
\end{equation}
Our goal is to generate an image $\hat{y} = G(x,p)$ which resembles ground-truth $y$. The first term of Eqn.~\ref{loss_gen} is the $L1$ loss function. $\Phi_k(.)$ denotes the feature maps of an image at $k$-{th} layer of a visual perception network. We use $VGG19$ \cite{simonyan2014very} network which is trained on ImageNet dataset. $\lambda_k$ is a hyperparameter which represents the importance of  $k$-{th} layer towards loss function. The second term in Eqn.~\ref{loss_gen} measures the perceptual similarity between an input image $y$ and an output image $\hat{y}$. We refer $L_{G}(G)$ as reconstruction loss.

In order to train discriminators, we also consider additional poses taken from different fashion item as shown in Fig.~\ref{proposed_model}. 
Let us denote these additional poses as $\tilde{p}$. With $\tilde{p}$ as input, the generator $G$ will produce new images $\hat{{y}}^\prime$.
$D_{I}$ discriminator aims to identify generated images as   `fake'. In order to learn the parameters of $D_{I}$, we adopt adversarial training as presented in \cite{goodfellow2014generative}. The loss function can be written as 
\begin{equation}
L_{D_{I}}(G, D_I ) = \mathbb{E} [ \log{D_I(y)} ] + \mathbb{E} [1 - \log{D_I(G(x , \tilde{p}))}]
\end{equation}
Similarly, $D_{P}$ discriminator distinguishes between real and fake by checking the consistency between given image and pose pair.  The loss function for $D_{P}$ can be written as
\begin{equation}
L_{D_{P}}(G, D_P ) = \mathbb{E} [ \log{D_P(\tilde{y},\tilde{p})} ] + \mathbb{E} [1 - \log{D_P(G(x , \tilde{p}),\tilde{p})} ]
\end{equation}
$\tilde{y}$ and $\tilde{p}$ represent image samples different from input image and corresponding pose map in training set respectively.
We formulate our full objective  as 
\begin{equation}
   G^{\star}, D^{\star}_I, D^{\star}_P = \arg \min_{G} \max_{D_1, D_2}  L_{G} + \alpha L_{D_{I}} + \beta L_{D_{P}}
    \label{all_loss}
\end{equation}
$\alpha$ and $\beta$ are the weights on the loss functions of two discriminators. 

\section{Experimental Results} 
In this section, we demonstrate our experimental results for generating photorealistic images of a person guided by target pose. 
We evaluate the proposed network on two datasets: DeepFashion \cite{liu2016deepfashion} and Market-1501 \cite{zheng2015scalable}. We show both qualitative and quantitative results and compare our approach against recent state-of-the-art pose-guided image generation methods.

\begin{table*}[h]
\begin{center}
\begin{tabular}{ccccccccc}
\hline \hline
&\multicolumn{2}{c}{} & \multicolumn{2}{c}{DeepFashion} & \multicolumn{2}{c}{} &\multicolumn{2}{c}{Market-1501}\\ [0.5ex]
  \cline{4-5}  \cline{8-9}
Methods & \multicolumn{2}{c}{} &  SSIM & IS & \multicolumn{2}{c}{} & SSIM & IS\\ [0.5ex]
\hline
Real Data  & \multicolumn{2}{c}{} & $1.000$ & $3.415$ & \multicolumn{2}{c}{} & $1.000$ & $3.678$\\ 
\hline 
pix2pix \cite{isola2017image} & \multicolumn{2}{c}{} & $0.646$ & $2.640$ & \multicolumn{2}{c}{} & $0.166$ & $2.289$\\ 
PG$^2$($G1+D$) \cite{ma2017pose} & \multicolumn{2}{c}{} & $0.761$ & $3.091$ & \multicolumn{2}{c}{} & $0.283$ & $3.490$\\ 
PG$^2$ \cite{ma2017pose} & \multicolumn{2}{c}{} & $0.762$ & $3.090$ & \multicolumn{2}{c}{} & $0.253$ & $3.460$\\ 
Varational U-Net \cite{esser2018variational} & \multicolumn{2}{c}{} & ${0.786}$ & $3.087$ & \multicolumn{2}{c}{} & $\textbf{0.353}$ & $3.214$ \\
\hline
$G+D_I$ & \multicolumn{2}{c}{} & $0.715$  & $3.053$ & \multicolumn{2}{c}{} & $0.216$ & $3.380$\\ 
$G+D_P$ & \multicolumn{2}{c}{} & $0.706$  & $2.778$ & \multicolumn{2}{c}{} & $0.197$ & $3.020$\\ 
 ($G+D_I+D_P$)-$\textbf{S}$ & \multicolumn{2}{c}{} & $0.782$  & $2.942$ & \multicolumn{2}{c}{} & $0.313$ & $3.390$\\ 
($G+D_I+D_P$)-$\textbf{M}$ & \multicolumn{2}{c}{} & $\textbf{0.789}$  & ${3.006}$ & \multicolumn{2}{c}{} & $0.344$ & ${3.291}$\\ [0.5ex]
 
\hline
 \end{tabular}
\end{center}
\caption{The table demonstrates the structural similarity (SSIM) and the inception score (IS) of our proposed model and other state-of-the-art methods on DeepFashion \protect\cite{liu2016deepfashion} and Market-1501 \protect\cite{zheng2015scalable} datasets. $(G + D_I + D_P)-\textbf{S}$ denotes proposed model with single input image, while $(G + D_I + D_P)-\textbf{M}$ indicates proposed model that takes multiple images as input.
}
\label{quant_res} 
\end{table*}

\subsection{Dataset}
In our experiment, DeepFashion \cite{liu2016deepfashion} and Market-1501 \cite{zheng2015scalable} datasets are used for evaluation.
We use In-shop Cloth Retrieval benchmark from DeepFashion dataset. DeepFashion includes multiple images of a person with different poses. This dataset contains $52,712$ in-shop
clothes images. We use the same training and testing set as presented in \cite{ma2017pose}. The resolution of an image is $256 \times 256$. 
We also demonstrate our experimentation on Market-1501 dataset. This dataset is very challenging due to variety of pose, illumination, and background.  It has $32,668$ images with  $128 \times 64$ resolution of $1,501$ persons captured from six different view points. For fair comparison, we follow PG$^2$ \cite{ma2017pose} for splitting training and testing sets. 

\begin{figure*}[!t]
\begin{center}
\begin{tabular}{c}
\rule{0pt}{0pt}
    \vspace{-1pt}\\ 
    \includegraphics[width=.72\linewidth]{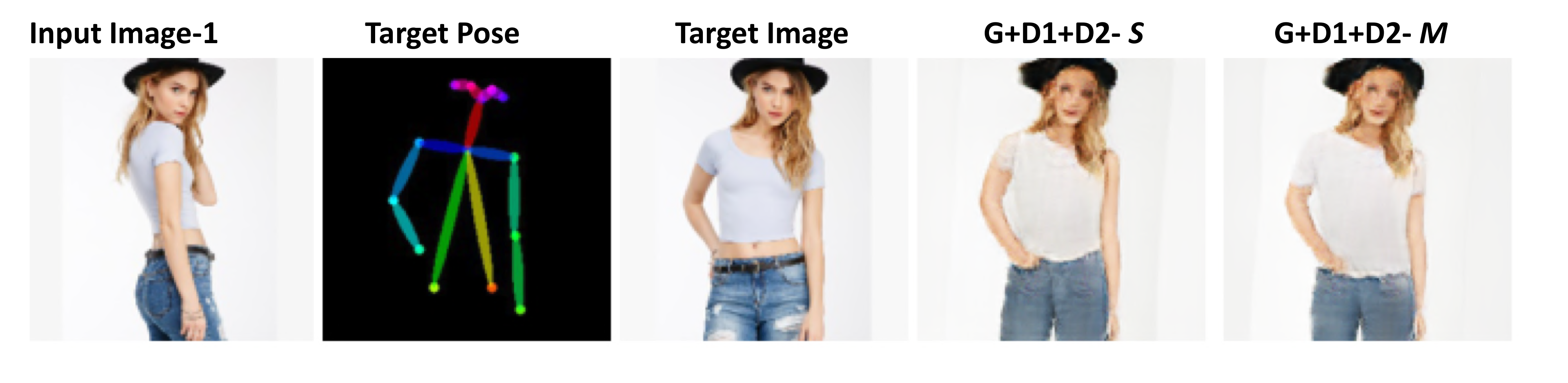} \vspace{-2 pt} \\
    \includegraphics[width=.72\linewidth]{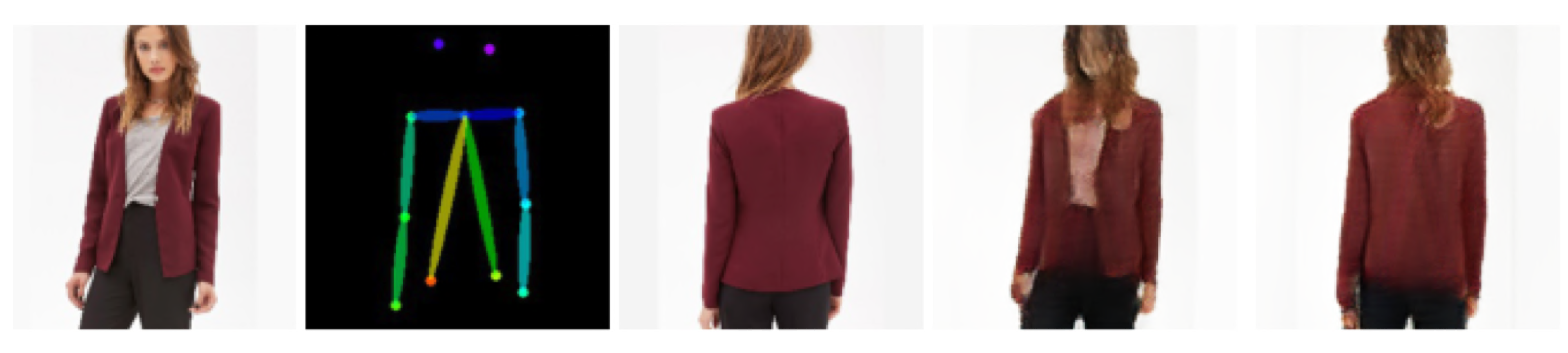} \vspace{-8 pt} \\
    \includegraphics[width=.72\linewidth]{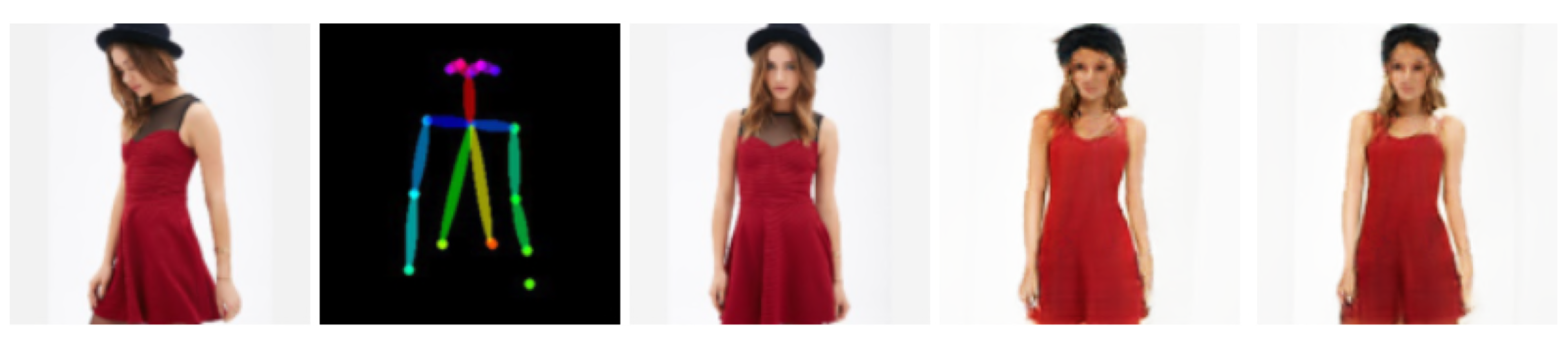} \vspace{-8 pt} \\
    \includegraphics[width=.72\linewidth]{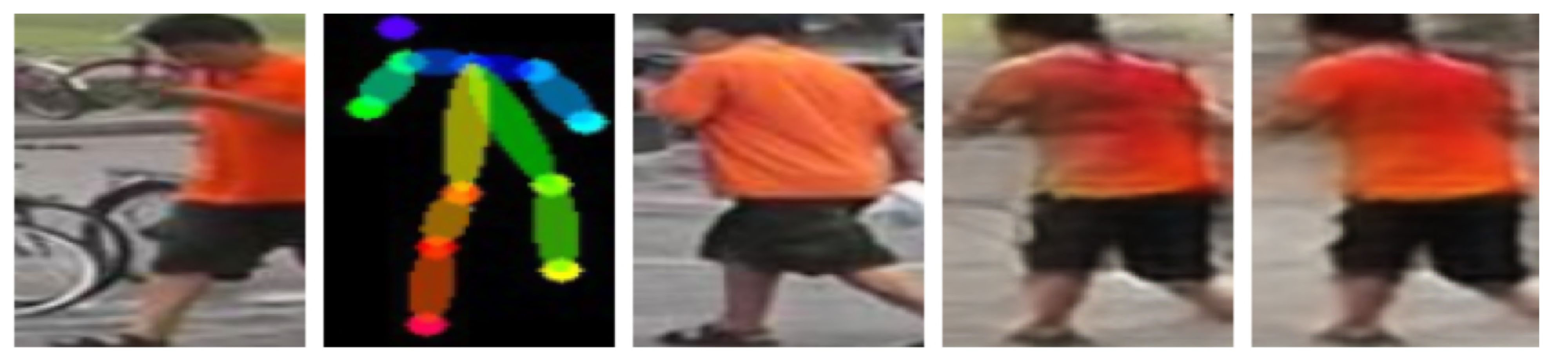} \vspace{-8 pt} \\
    \includegraphics[width=.72\linewidth]{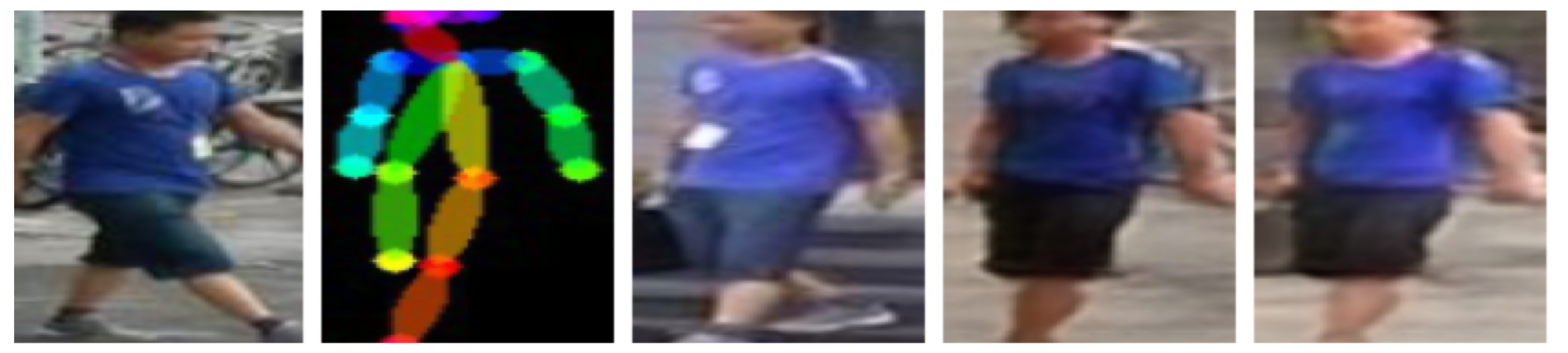} \vspace{-8 pt} \\
    \includegraphics[width=.72\linewidth]{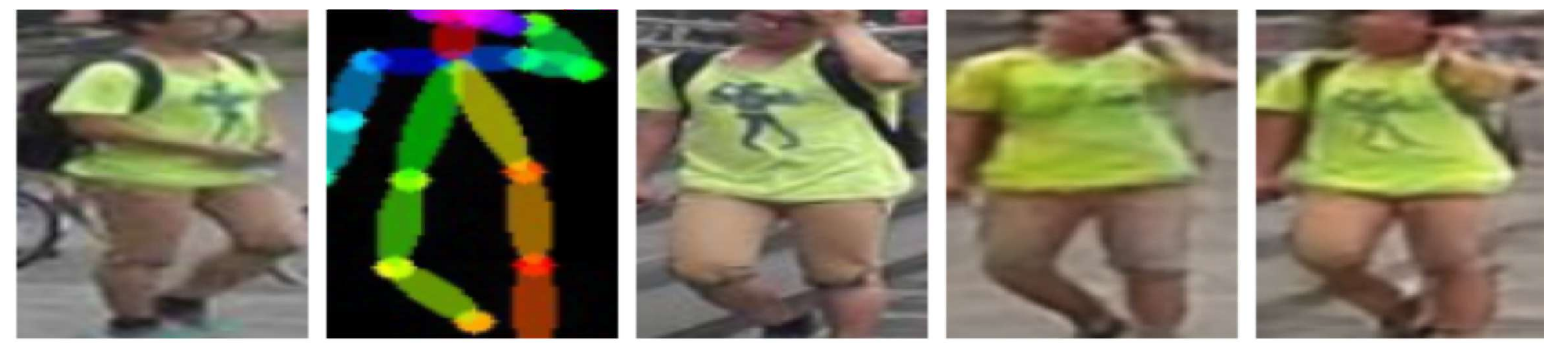} \vspace{-8 pt} \\
 \end{tabular}

\end{center}
  \caption{\small The figure shows some image generation results by our proposed framework. First and second columns represent input images. Third and fourth columns show the target pose and target image respectively. Synthetic  images produced by $G+D_I+D_P-S$ and $G+D_I+D_P-M$ methods are shown in fifth and sixth columns respectively. Please see Sec.~\ref{qual_analysis} for more details. }
\label{qual_res} 
\end{figure*}

\subsection{Implementation Details}
Our U-Net encoder and decoder follow the network architecture as presented in \cite{zhu2017unpaired}. 
The network
contains two stride-2 convolution layers and several residual blocks, and   two  fractionally-strided  convolutions  with  stride \(\frac{1}{2}\).  Each layer of the image encoder only contains convolutional residual blocks. 
For DeepFashion dataset, we 
use $6$ residual blocks. 
In order to train two discriminators $D_{I}$ and $D_{P}$, we adopt the training procedure as PatchGAN \cite{isola2017image}. The discriminator uses $70 \times 70$ patch and averages all scores by sliding across the image to determine the final output. This allows us to capture high frequency structures. 
To optimize the network parameters, we use Adam optimizer \cite{kingma2014adam} with
$\beta_{1} = 0.5$ and $\beta_2 = 0.999$. We use batch size of $1$, with initial learning rate $1e^{-4}$, decay $0.5$ every $50$ epochs. 
Here, batch size represents one SKU which includes multiple images of a fashion item ranging from $2$ to $5$. These images are further used for data augmentation. We randomly crop images and flip the image left-right for data augmentation. We also randomly rotate images to increase the training set.




\subsection{Quantitative Results}

In order to evaluate our proposed model, we consider two metrics to measure the quality of image synthesis. We utilize Structural Similarity (SSIM) \cite{wang2004image} and the Inception Score (IS) \cite{salimans2016improved} as evaluation criteria.
Table.~\ref{quant_res} shows the  quantitative results on DeepFashion and Market-1501 datasets. Next, we will compare against some baseline and state-of-the-art methods. 

 

\subsubsection{Impact of Two Discriminators}
Unlike most of the pose guided image generation methods \cite{ma2017pose,esser2018variational}, we take advantage of adversarial training with two discriminators -  $D_I$ and  $D_P$. In order to analyze the effect of these two discriminators we remove one discriminator at a time, and evaluate the performance. If we remove $D_I$ from the network, the loss function $L_{D_I}$ in Eqn.~\ref{all_loss} does not have any impact. In other words, the mapping function $D_I$ in Eqn.~\ref{all_loss} has no contribution in the network. We denote this model as $G + D_I$. 
To verify the effectiveness of the two discriminators,
we pick DeepFashion dataset to run  the framework with different configurations. Furthermore, we provide the results of our proposed model with two discriminators on Market1501 dataset as shown in Table.~\ref{quant_res}. 
As can be seen in Table.~\ref{quant_res}, after removal of  $D_I$, both SSIM and IS scores have been significantly dropped by $10.52\%$ and $7.58\%$ respectively compared with $(G + D_I + D_P)-\textbf{M}$ on DeepFashion dataset. 
For Market-1501 dataset,  $(G + D_I + D_P)-\textbf{M}$ model outperforms $G + D_I$ model by large margin both in IS ($8.23\%$) and SSIM ($46.96\%$) scores.
Since $D_I$ distinguishes whether an image is real or generated, $G + D_P$ model can not generate photorealistic images with high SSIM and IS score.

Similarly, we refer the removal of $D_P$ discriminator from the proposed architecture as $G + D_P$. $D_P$ helps the model to generate photorealistic images of a person with target pose by comparing between real image-pose pair and generated image-pose pair. From Table.~\ref{quant_res}, we can see that the $G + D_I$ achieves $3.059$ and $0.715$ in IS and SSIM scores respectively.  We observe a large drop ($~9.38\%$) in SSIM score with compared to proposed model $(G + D_I + D_P)-\textbf{M}$. The SSIM score can be improved by exploiting  $D_P$ discriminator as shown in Table.~\ref{quant_res}. 

\subsubsection{Effect of Using Multiple Images}

In our proposed architecture, we exploit multiple photos of a same fashion item to extract visual-semantic features. 
During the training process, we allow bi-directional convolutional LSTM (bC-LSTM) network to learn common attributes by observing the transition between multiple images. These attributes or visual-semantic features are utilized to generate photorealistic images. The proposed model is also capable of taking single image as input. 
Table.~\ref{quant_res} shows its SSIM and IS score on  DeepFashion and Market-1501. 

Multi-Image mode outperforms Single-Image model by large margin ($0.89\%$ on DeepFashion and $9.90\%$ Market-1501  datasets) in terms of SSIM score. Multi-Image model also achieves better IS score compared to Single-Image model on Deepfashion dataset. 
From Table.~\ref{quant_res}, we can conclude that bC-LSTM in the generator learns visual-semantic contextual details by exploiting multiple images as input. 

\subsubsection{Compare against State-of-the-art}
In this section, we compare our proposed model against other state-of-the-art deep generative models. We choose some recent works - PG$^2$ \cite{ma2017pose},  PG$^2(G1+D)$ \cite{ma2017pose}, pix2pix \cite{isola2017image}, and variational U-Net \cite{esser2018variational} to evaluate the performance of our proposed model.  
From the Table.~\ref{quant_res}, we can see that the proposed method 
achieves comparable results with other exiting works in terms of SSIM score.
To measure the quality of image generation process, we also compare the proposed approach with other state-of-the-art models by exploiting IS score as shown in Table.~\ref{quant_res}. The proposed model
outperforms PG$^2$ \cite{ma2017pose},  PG$^2(G1+D)$ \cite{ma2017pose}, pix2pix \cite{isola2017image}, and variational U-Net \cite{esser2018variational} by large margin in  IS score on DeepFashion and datasets. 
Market-1501 \cite{zheng2015scalable} dataset has images with various background which becomes very difficult to predict as there is no information of background in target image from the input.
Our model is able to generate photorealistic images with high SSIM and IS score on Market-1501 dataset as presented in Table.~\ref{quant_res}. Our model get state-of-the-art performance in terms of Inception Score, which indicates our model not only generate realistic images, but also output a high diversity of images, and with a lower probability to mode collapse. As for SSIM, we achieve improvement to \cite{ma2017pose} and comparable results with \cite{esser2018variational}.

\subsection{Qualitative Analysis}
\label{qual_analysis}
Given an image or multiple images of a fashion item along with target pose, our proposed model is able to transfer a person's current pose to intended pose. 
Fig.~\ref{qual_res} illustrates some image synthesis results produced
by Single-Image model and Multi-Image model. The examples are taken from DeepFashion and Market-1501 datasets. 
$(G + D_I + D_P)-\textbf{S}$ takes a single image as shown in $1^{st}$ column of Fig.~\ref{qual_res} and a pose map as input. The synthesis results by $(G + D_I + D_P)-\textbf{S}$ are shown in $4^{th}$ column.
In Fig.~\ref{qual_res}, 
we also show the generation results obtained from  Multi-Image model. 
$(G + D_I + D_P)-\textbf{M}$ model takes multiple images as input. For convenience, we only show one image in Fig.~\ref{qual_res}.
$5^{th}$ column exhibits the generated images created by $(G + D_I + D_P)-\textbf{M}$. 
From  Fig.~\ref{qual_res}, we can see  the high resemblance between the synthetic ($4^{th}$ and $5^{th}$ columns) and target ground truth images ($3^{rd}$ column) as shown in the figure. Furthermore, the proposed model is also able to predict reasonable face details such as mouth, eyes and nose of a person as illustrated in Fig.~\ref{qual_res}.  



\section{Conclusion}

In this paper, we present a novel generative model to produce photorealistic images of a person change to a target pose. We utilize convolutional LSTM, and U-Net architecture to develop the generator, in which we 1) exploit multiple images of the same person in order to learn semantic visual context from the convolutional LSTM network; 2) apply a U-Net encoder to learn the appearance/geometrical information; and 3) use a U-Net decoder to generate an image by exploiting visual and appearance context. In order to better guide the image generation process, we apply two discriminators specifically designed for image authenticity and pose consistency. Our experimental results show that the proposed model can produce high-quality images both in qualitative and quantitative measures. As future direction, we will explore the usage of visual and appearance context for human parsing.

\vspace{3mm}

\bibliographystyle{ACM-Reference-Format}
\bibliography{main}

\end{document}